# Training of Deep Learning Neuro-Skin Neural Network

**Mehrdad S. Dizaji**

**Abstract**

In this brief paper, a learning algorithm is developed for Deep Learning Neuro-Skin Neural Network to improve their learning properties. Neuroskin is a new type of neural network presented recently by the authors. It is comprised of a cellular membrane which has a neuron attached to each cell. The neuron is the cell's nucleus. A neuroskin is modelled using finite elements. Each element of the finite element represents a cell. Each cell's neuron has dendritic fibers which connects it to the nodes of the cell. On the other hand, its axon is connected to the nodes of a number of different neurons. The neuroskin is trained to contract upon receiving an input. The learning takes place during updating iterations using sensitivity analysis. It is shown that while the neuroskin can not present the desirable response, it improves gradually to the desired level.

*Key words* - Neuro-skin, L-BFGS-B Algorithm, Training, Neural Networks, Finite Element

## 1. NEURO-SKINS OR NERO-MEMBRANES (NMS)

In the previous paper, neuroskins are presented and their response characteristics are studied [1]. Neuroskins are an improved version of Dynamic Plastic Neural Networks (DPNN)s which have recently been introduced by the authors [1-2]. To get more information about the neuroskin neural networks and their characteristics, reader can refer to the previous research [1]. The neuro-membrane utilizing in this paper is a two-dimensional plate which has been studied by the authors in their previous papers on DPCNNs [1] – [26]. They have used this problem in all their studies so that the results and characteristics of the different continuous neural network models could be compared. Figure 1 shows the plate which is 500 mm by 1000 mm. It is unrestricted on three of its edges but restricted by 11 simple supports on its fourth edge [1].

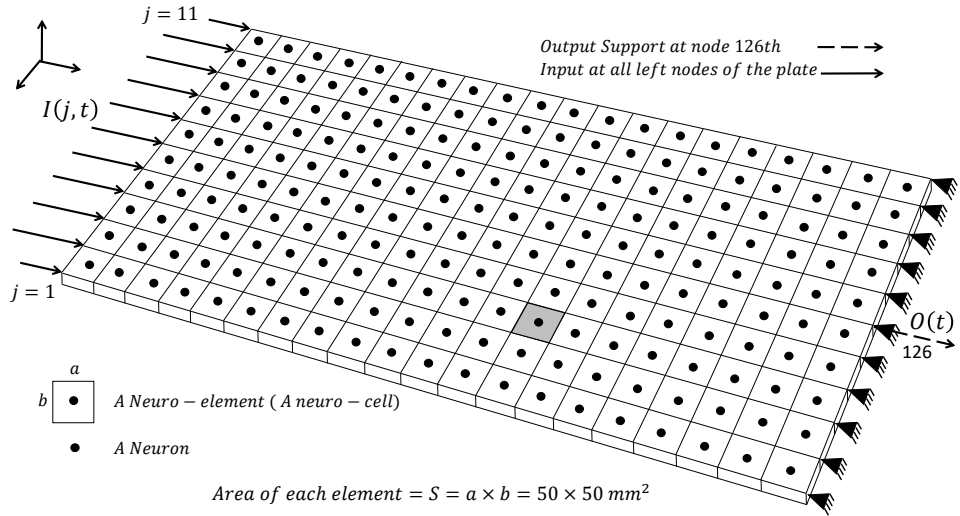

**Figure 1.** Finite-Neuro-Elements mesh studied in this paper has 10×20 square Neuro-Elements (cells), each containing a neuron. The size of all the square elements is 50 mm: inputs, output and elements with their neurons,

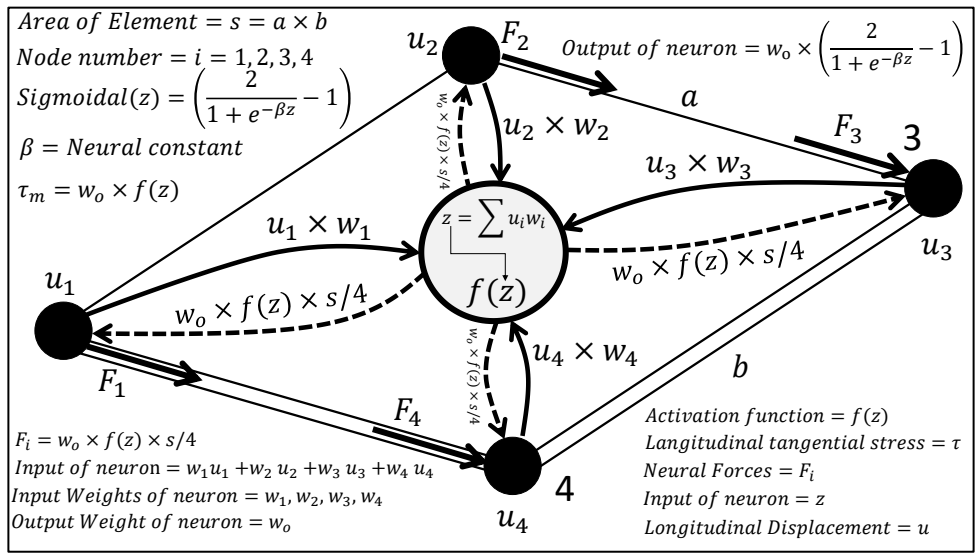

**Figure 2.** Isometric view of a 4-node rectangular Neuro-Element with its neuron. For the square element used in this paper, $a=b$. The neuron shown here is sigmoidal but can be any type neuron, as is discussed later in the paper.

Figure 2 shows an isometric view of a neuro-element used in this paper, with all the information about its neuron, input connection weights and its output weight. The figure also explains how the output of the neuro-element is calculated. The horizontal displacements of nodes are multiplied by their corresponding input weights. The result is delivered to the neuron's processor where $f(z)$, the neuron's activation function, receives the input $z$ and issues its output $f(z)$ which is a value in [-1, +1]. The output is first multiplied by the neuron's output weight ($w_o$), then multiplied by the neuro-element's area to determine the total traction force of the element. The traction is divided by 4 and is applied to each node of the neuro-element to be assembled later as a part of the total nodal force. The behaviour and characteristics of a neuro-membrane depends on the type of the activation function incorporated in it. This issue is studied in the following sections where different types of activation functions are considered for the example problem in this paper and the response of the neuro-membrane has been studied. Also, the adaptivity of the neuro-membranes to learn a dynamic input-output relationship, similar to multi layer feed forward neural networks, is investigated briefly without aiming to provide final details [1].

## 2. Training Algorithm Neuro-Skins or Nero-Membranes (NMs)

The adaptive parameters of the Neuro-Skin neural network are its connection weights. Generally, the error of the Neuro-Skin can be expressed in terms of the adaptive parameters. The mean Squared Error (MSE) of the Neuro-Skin Neural Network is:

$$MSE(w) = \frac{1}{N}\sum_{i=1}^{N}(\hat{y}(t_i) - y(t_i))^2 \tag{1}$$

Where *w*, *N* and $\hat{y}(t_i)$ are connection weights, number of training pairs and output of the Neuro-Skin Neural Network, respectively. The L-BFGS-B method is successfully applied for training process of the Neuro-Skin Neural Network. In the Figure 3, training procedure of the Neuro-Skin Neural Network is shown schematically. The design variables are selected as an output to each neuron which are tuned during learning procedure. The results for convergence of the design variables to their target values indicates the capability of the introduced training algorithm.

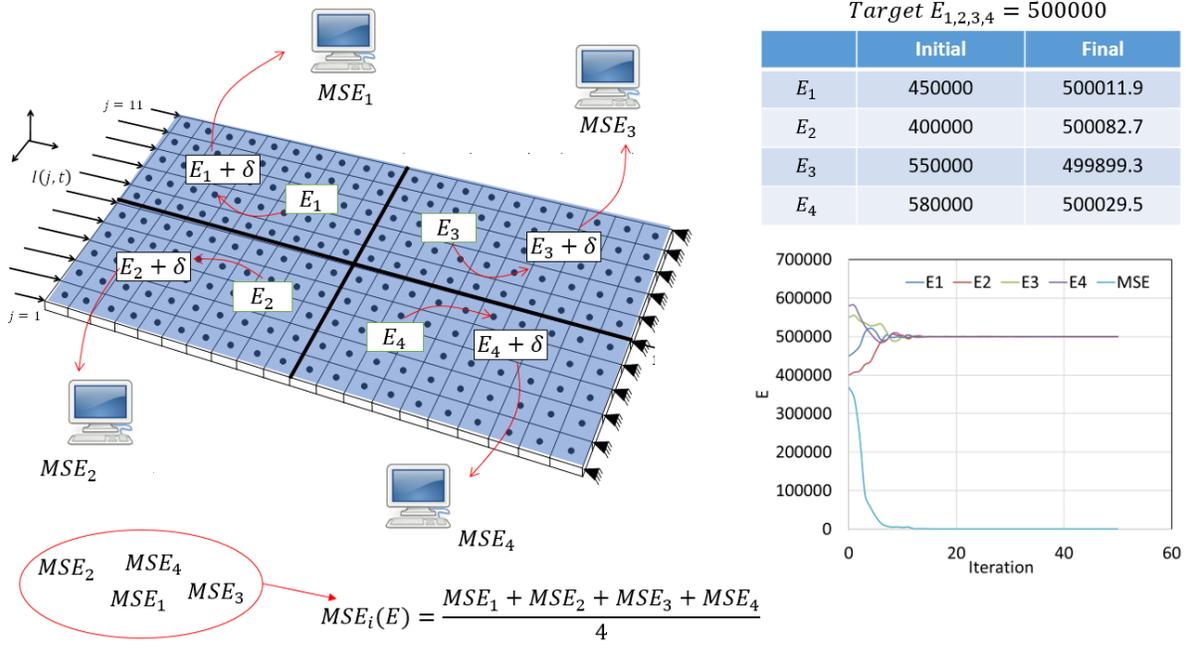

**Figure 3**. Training algorithm of the Deep Learning Neuro-Skin Neural Network schematically

## 3. CONCLUSION

A new algorithm based on sensitivity analysis for the process of training of Neuro-skin was introduced. The neuro-skin is adaptive and has the capability of providing desirable response to inputs intelligently. Hence, after training, it responds as a smart medium and this way, the neuro-skin can be considered as a new type of neural network, called Deep Learning Neuro Skin Neural Networks, with adaptability and learning capability. Training of the neuro-skin to represent a desired output was an essential part of the study. The paper shows that the defined training algorithm of the neuro-skin by dynamic finite element method is a suitable approach.

## 4. APPENDIX

```python
from multiprocessing import Pool,cpu_count
import subprocess
import os
import shutil
import fileinput
import numpy as np
from scipy.optimize import fmin_bfgs,fmin,fmin_l_bfgs_b

ansys_dir='C:\\Program Files\\ANSYS Inc\\v171\\ansys\\bin\\winx64'

target=np.loadtxt('output.out')

def run(num,x):
    if not os.path.exists(str(int(num))):
        os.makedirs(str(int(num)))
        current_dir=os.getcwd()
        current_dir +='\\'+str(int(num))
        shutil.copy('aaa.txt',current_dir)
        shutil.copy('bbb.txt',current_dir)
        shutil.copy('fne.txt',current_dir)
    else:
        current_dir=os.getcwd()
        current_dir +='\\'+str(int(num))

    with open(current_dir+'\\aaa.txt','w') as f:
        k=200/len(x)
        for i in range(len(x)):
            for j in range(k):
                m=1+i*k+j
                f.write('v'+str(int(m))+'='+str(x[i])+'\n')

    command=ansys_dir+'\\ansys171 -b -np 1 -dir "'+current_dir +'" -i fne.txt -o analysis.o'
    os.chdir(current_dir)
    subprocess.call(command)
    return np.loadtxt('output.out')

def run_ex(x):
    return run(*x)
```

```python
def func(x):
    p=Pool(len(x)+1)
    input_list = [(0,x)]
    delta = 1.0e-2
    for i in range(len(x)):
        copy = np.array(x)
        copy[i] += delta
        input_list.append((i+1,copy))
    out=p.map(run_ex,input_list)
    f=np.sqrt(np.mean((out[0]-target)**2))
    g=[]
    for i in range(len(x)):
        f1=np.sqrt(np.mean((out[i+1]-target)**2))
        g.append((f1-f)/delta)
    return f,np.array([g])

def evaluate(xx):
    p=Pool(cpu_count())
    input_list=[]
    for i in range(len(xx)):
        input_list.append([i,xx[i]])
    out=p.map(run_ex,input_list)
    f=[]
    for i in range(len(out)):
        f.append(np.sqrt(np.mean((out[i]-target)**2)))
    return f

def call_back(xk):
    with open('result.txt','a') as f:
        st=''
        for xx in xk: st+=str(xx)+','
        st=st[:-1]+'\n'
        f.write(st)

if __name__ == '__main__':
    if os.path.isfile('result.txt'):
        os.remove('result.txt')
    xopt,fopt,d= fmin_l_bfgs_b(func, x0=[450000.0],bounds=[[400000,550000]],approx_grad=False,epsilon=0.001,...
    factr=1.0e12,maxfun=100,maxiter=5,iprint=1,callback=call_back)
    #f=open("final.txt",'w')
    #f.write(str(xopt)+','+str(fopt))
    #f.close()
    print (xopt,fopt)
    x=np.loadtxt('result.txt',delimiter=',')
    x=np.reshape(x,(len(x),-1))
    x=x.tolist()
    mse=evaluate(x)
    out=np.column_stack((x,mse))
    np.savetxt('result.txt',out,delimiter=',')
```